\documentclass{article}

\PassOptionsToPackage{numbers, compress}{natbib}
\usepackage[preprint]{neurips_2019}

\usepackage[utf8]{inputenc} 
\usepackage[T1]{fontenc}    
\usepackage{hyperref}       
\usepackage{url}            
\usepackage{booktabs}       
\usepackage{amsfonts}       
\usepackage{nicefrac}       
\usepackage{microtype}      
\usepackage{algorithmic}
\usepackage{algorithm2e}
\usepackage{amsmath}
\usepackage{amssymb}
\usepackage{graphicx}
\usepackage{caption}
\usepackage{subcaption}

\graphicspath{{images/}}

\title{Cost-Sensitive Feature-Value Acquisition Using Feature Relevance}

\author{%
  Kimmo Kärkkäinen \\
  University of California, Los Angeles \\
  \texttt{kimmo@cs.ucla.edu} \\
  \And
  Mohammad Kachuee \\
  University of California, Los Angeles \\
  \texttt{mkachuee@cs.ucla.edu} \\
  \And
  Orpaz Goldstein \\
  University of California, Los Angeles \\
  \texttt{orpgol@cs.ucla.edu} \\
  \And
  Majid Sarrafzadeh \\
  University of California, Los Angeles \\
  \texttt{majid@cs.ucla.edu}
}

\begin{document}

\maketitle

\begin{abstract}
In many real-world machine learning problems, feature values are not readily available. To make predictions, some of the missing features have to be acquired, which can incur a cost in money, computational time, or human time, depending on the problem domain. This leads us to the problem of choosing which features to use at the prediction time. The chosen features should increase the prediction accuracy for a low cost, but determining which features will do that is challenging. The choice should take into account the previously acquired feature values as well as the feature costs. This paper proposes a novel approach to address this problem. The proposed approach chooses the most useful features adaptively based on how relevant they are for the prediction task as well as what the corresponding feature costs are. Our approach uses a generic neural network architecture, which is suitable for a wide range of problems. We evaluate our approach on three cost-sensitive datasets, including Yahoo! Learning to Rank Competition dataset as well as two health datasets. We show that our approach achieves high accuracy with a lower cost than the current state-of-the-art approaches.

\end{abstract}

\section{Introduction}

Traditionally, research on machine learning algorithms has focused on achieving accurate predictions on fully available feature sets. However, in the real world, data is often incomplete and acquiring additional feature values will incur a cost. For example, when making a medical diagnosis, a doctor examines the patient and determines what further information is needed to make a diagnosis. The next question to ask has to be chosen out of a vast number of possibilities, but the number of potentially useful follow-up questions can be narrowed down with the answers received for each question. Once the doctor has acquired enough information on the patient, they make a diagnosis and choose the appropriate treatment. Acquiring information could mean e.g. performing medical tests (blood tests, imaging, etc) or asking the patient for more subjective information on their symptoms. In this case, there are monetary costs for each test as well as for the doctor's time, and these costs can be very different. Asking the patient for information is fast and cheap, whereas performing medical tests can have a high cost.

Money is not the only type of cost that needs to be taken into account. Medical tests can have negative side-effects or they can cause patient discomfort. In the field of mobile health, acquiring data can have an impact on the mobile device's battery life. Individuals might also be uncomfortable disclosing specific information, in which case there is a privacy cost for acquiring data. There could even be a cost based on how much human time is required to acquire a feature value. The approaches proposed in this paper can be used with any type of cost, as long as the costs are quantifiable on a linear scale.

There is a wide range of existing research on minimizing the costs associated with machine learning. We categorize these approaches into four different categories. The first category is feature selection, where the number of features is minimized in the training phase so that the cost becomes lower without affecting the prediction accuracy too much \citep{Chandrashekar2014AMethods}. This approach, however, does not take into account that some inputs are easier to make predictions on than others, so they should have a lower cost as well. The second category is cascade algorithms, where at each step a decision is made to either acquire the next feature or stop and make a prediction \citep{Xu2012TheBudgets,Chen2012ClassifierCost,Trapeznikov2013Multi-stageDesign,Wang2014AnBudgets}. In this approach, the order of features is decided in the training phase, so they have only a limited ability to adapt to newly acquired information. The third category is trees of classifiers, in which the new information affects which feature is acquired next \citep{xu2013cost,Kusner2014Feature-costClassifiers,Wang2015EfficientPrediction}. However, the number of classifiers can become considerable if there is a large number of features. The fourth category is fully adaptive algorithms, which can choose any feature based on what is most valuable in the current situation \citep{Chen2015ComputerProbability,Early2016TestFOCUS,kachuee2018dynamic,kachuee2019opportunistic}. There are only a few algorithms in this category, and they tend to have a high computational cost.

The algorithms proposed in this paper belong to the last category, where any unknown feature can be chosen for acquisition. Furthermore, the feature acquisition decisions are based on the knowledge of the existing features. We have chosen to use neural networks as our predictive model, as they have been shown to perform well across many domains. We use generic neural network architectures to demonstrate that our proposed algorithms do not depend on domain-specific structures. To choose which features to acquire next, we derive our approach from the recent research on relevance propagation \citep{Bach2015OnPropagation,Montavon2017ExplainingDecomposition}. The focus of that line of research has been to explain why a neural network made a specific prediction, but we show that a similar approach can give us valuable information on the importance of missing feature values as well. To the best of our knowledge, relevance propagation has not been used in active feature acquisition previously. The benefit of our approach is the ease of training a model and the relatively low computational cost in both the training and testing phases.

In this paper, we propose two algorithms for active feature acquisition. The first algorithm achieves high accuracies at a low computational cost. The second algorithm improves the results of the first one at the expense of a slightly higher computational cost. Our main contributions can be summarized as follows:

\begin{itemize}
  \item We demonstrate how relevance propagation can be utilized for the active feature acquisition problem.
  \item We develop an efficient algorithm that uses feature relevance information for feature acquisition. We show through experiments that feature relevances can give us valuable information on the usefulness of missing features with a very low computational cost.
  \item Based on the first algorithm, we develop a second algorithm that provides better results at a slightly increased computational cost.
  \item We compare our results with other state-of-the-art algorithms and show that our approach achieves a high accuracy with a lower cost on three realistic datasets.
\end{itemize}

\section{Relevance Propagation}
\label{section:relprop}

The approach proposed in this paper is related to the recent research on layer-wise relevance propagation (LRP) \citep{Bach2015OnPropagation}. The goal of LRP is to show how large an impact each of the input features had on the prediction, which is a non-trivial task on neural networks due to their non-linearities. To simplify this problem, relevances are propagated back to the input layer one layer at a time following a few specific rules. First, the relevance of the output layer is set to the predicted output value:

\begin{equation}
R_{out} = f(\mathbf{x})\,,
\end{equation}

where $f(\mathbf{x})$ is the output of our neural network with an input vector $\mathbf{x}$. 

Second, each layer should also have the same total relevance:

\begin{equation}
R_{out} = \sum_{i=1}^{N_l} R_i^l = \sum_{i=1}^{N_{l-1}} R_i^{l-1} = \ldots = \sum_{i=1}^{N_{1}} R_i^{1}\,,
\end{equation}

where $R_i^l$ is the output relevance of neuron $i$ belonging to layer $l$ and $N_l$ is the number of neurons in layer $l$. This constraint guarantees that all of the relevance on the higher layer will be distributed to the neurons on the lower layer without adding or removing any relevance between layers. However, the distribution of the relevance within the layers can differ. Doing this at every layer means that the total relevance on the input layer is equal to the relevance on the output layer. Furthermore, the output relevance of a neuron should be equal to the sum of relevances of lower-level neurons that are directly connected to it:

\begin{equation}
R_i = \sum_j R_{i \leftarrow j}\,,
\end{equation}

where $R_{i \leftarrow j}$ is the amount of relevance attributed from a higher-level neuron $j$ to neuron $i$. These rules are fulfilled when a neuron distributes its output relevance entirely to its inputs according to specific rules. 

There are multiple approaches for distributing the relevances to the lower layer. The approach proposed by \citet{Bach2015OnPropagation} is to distribute the relevance of one neuron to the input neurons in the same proportion as the input values received from each neuron:

\begin{equation}
R_{i \leftarrow j} = \frac{z_{ij}}{\sum_{j'} z_{ij'}} \cdot R_j\,,
\end{equation}

in which $z_{ij} = x_i w_{ij}$, $x_i$ is an input value for the neuron, and $w_{ij}$ in the corresponding weight. As this rule might not be numerically stable with small activation values, they present a few alternative approaches to mitigate the problem, but the overall idea remains the same.

\citet{Montavon2017ExplainingDecomposition} proposed another way to propagate relevances while still following the other rules defined by \citet{Bach2015OnPropagation}. Their approach, called Deep Taylor Decomposition, approximates each neuron using a first-order Taylor series approximation to determine how much relevance should be propagated to each neuron input. Using this approach, they derive three rules for propagating the relevance based on the neuron's input domain. When the neuron has unconstrained input, the propagation rule becomes:

\begin{equation}
R_i = \sum_j \frac{w_{ij}^2}{\sum_{i'} w_{i'j}^2} R_j\,.
\end{equation}

If the input is constrained to have only positive values, as is the case when the previous layer contains rectified linear units (ReLU), the propagation rule is:

\begin{equation}
R_i = \sum_j \frac{z_{ij}^+}{\sum_{i'} z_{i'j}^+} R_j\,,
\end{equation}

where $z_{ij}^+=x_i w_{ij}^+$, and $w_{ij}^+$ is the positive part of the weight $w_{ij}$. The positive part of a vector is defined as a vector that has all the negative values replaced by zeroes. Finally, if the neuron input is known to have specific upper ($h_i$) and lower ($l_i$) bounds, as is often the case in the first layer, the rule becomes:

\begin{equation}
R_i = \sum_j \frac{z_{ij} - l_i w_{ij}^+ - h_i w_{ij}^-}{\sum_{i'} z_{i'j} - l_{i'} w_{i'j}^+ - h_{i'} w_{i'j}^-} R_j\,.
\end{equation}

Full derivations of these rules can be found in \citep{Montavon2017ExplainingDecomposition}.

\section{Our Approach}

\subsection{Problem Definition}
Let us consider a case where a feature vector $\textbf{x}_{\mathit{full}}$ has all feature values available, but only part of these values are known to us at any time. We know the value of $\textbf{x}_{\mathit{partial}} = \textbf{x}_{\mathit{full}} \odot \textbf{k}^t$, where $\textbf{k}^t$ is an indicator vector containing values 1 or 0 depending on whether the corresponding feature value is known to us at time $t$. The number of known features at time $t$ is therefore $|\textbf{k}^t|_0$. To acquire a new feature, we have to pay feature acquisition cost $c_i$, which can be different for each feature $i$. 

The feature acquisition algorithm will then take the following steps. First, the algorithm should determine the most valuable feature using the knowledge of the acquired feature values, the current prediction, and feature costs. This feature value is then requested from an oracle that is able to give any value for a cost, and the value is added to the feature vector of known values $\textbf{x}_{partial}$. The cumulative cost at each time step is therefore $ \sum_i c_i k_{i}^t $. This process is repeated until a stopping condition is reached. Possible stopping conditions can include e.g. the model certainty reaching a predefined level, the total cost becoming too high, or all of the feature values having been acquired. The appropriate stopping condition should be decided based on the problem domain. For example, in the medical domain it might be more appropriate to stop only when the model certainty is high enough, even if it leads to a higher cost. In less critical domains, minimizing the total cost might be more important.


Our task is then to define a value function $value(\textbf{x}_{partial}, \hat{\textbf{y}}, $\textbf{c}), which uses the currently known features, the current prediction, and the feature costs to determine the value of acquiring each unknown feature. This function should give the highest value for features that increase the prediction accuracy the most while having a low cost.

In this paper, we propose two methods for cost-sensitive feature acquisition where relevance propagation is used to determine feature informativeness. On a high level, the goal of these algorithms is to define which unknown feature is the most important for our current prediction at any given time. This importance should reflect our current knowledge of the available feature values. Our first method propagates the predicted value directly similar to how the earlier relevance propagation papers have done. Our second method propagates relevance through each output node separately to determine which feature is the most important over any potential class.

\subsection{Direct Propagation}
\label{sec:dirprop}

\begin{figure}[t]
\includegraphics[width=0.6\linewidth]{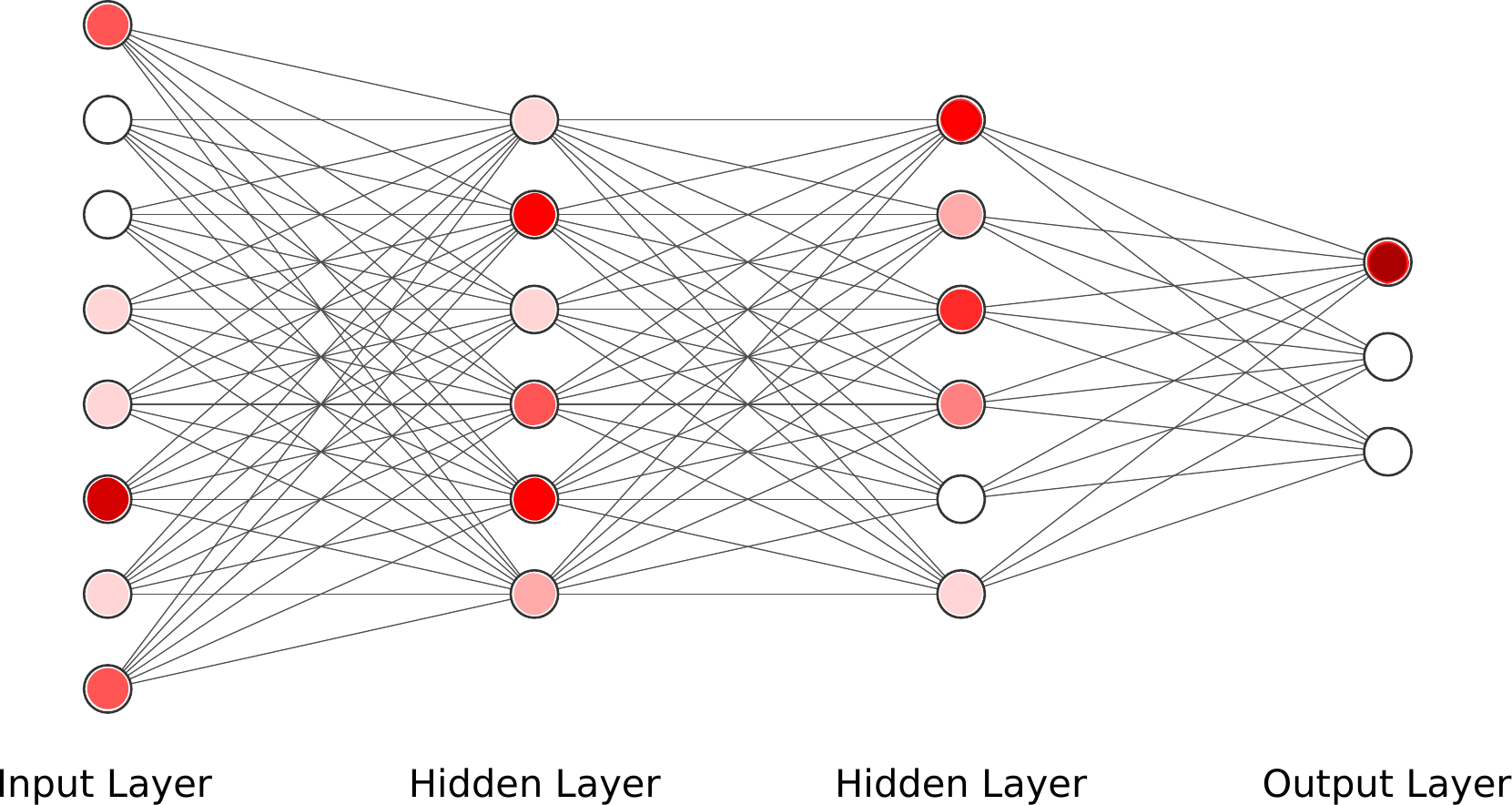}
\centering
\caption{Relevance is propagated from the output layer (on the right) to the input layer (on the left). Darker color indicates a higher relevance.}
\label{fig:nn}
\end{figure}

We start by training a neural network model $M$ with fully available features. Having fully available features is not a strict requirement but we choose to use them to make the results comparable with the existing research in this field. The model $M$ takes an input vector $\textbf{x} \in \mathbb{R}^m$ and produces a prediction $\hat{\textbf{y}} \in \mathbb{R}^n$. Our algorithm receives an input vector $\textbf{x}_{\mathit{partial}}$, which has no known values initially. To keep track of which feature-values have been acquired already, we introduce an indicator vector $\textbf{k} \in \mathbb{R}^m$, which has value 1 if the corresponding feature-value has been acquired and value 0 otherwise. The goal is then to request these feature values one-by-one until a prediction can be made. First, we fill in the missing values of the given input vector $\textbf{x}_{\mathit{partial}}$:

\begin{equation}
\textbf{x}_{\mathit{filled}} \gets \textbf{k} \odot \textbf{x}_{\mathit{partial}} + (\textbf{1}-\textbf{k}) \odot \mathbb{E}[\textbf{x}]\,,
\end{equation}

, where $\odot$ is the Hadamard product. The true expected value of $\textbf{x}$ is not known, so we estimate it using the training data. This gives us a feature vector where the missing values have been replaced with the expectation for those values. We then propagate $\textbf{x}_{\mathit{filled}}$ forward through our model $M$ to acquire a prediction $\hat{\textbf{y}}$. This prediction is then used as the starting point for relevance propagation. Using the Deep Taylor Decomposition propagation rules described in Section~\ref{section:relprop}, $\hat{\textbf{y}}$ is propagated backward to the input nodes. This gives us relevances $r_i$ for each input node, which is shown in Figure~\ref{fig:nn}.

Next, we will introduce an adjusted relevance score to take into account the constraints that our problem has. First, we need to notice that the relevances can have either positive or negative values, depending on whether the corresponding feature increased or decreased the predicted value. High impact in either direction can be important, so we will use the absolute value. Next, we need to take into account the feature costs. Two features could provide the same amount of information but have a vastly different cost, so we will normalize the earlier value using the corresponding feature cost. Finally, we are only interested in the unknown features, which leads us to the adjusted relevance $\hat{\textbf{r}}$ where individual values are defined as:

\begin{equation}
\hat{r}_i \gets (1-k_i) \cdot \frac{| r_i |}{c_i}\,.
\end{equation}

This defines all known feature values to have zero relevance, whereas unknown feature values have a relevance value depending on the magnitude of their original relevance as well as the associated feature acquisition cost. The relevance of known features is set to zero to avoid acquiring the same features multiple times. Finally, we acquire the feature value $x_j$, where $j = argmax(\hat{\mathbf{r}})$. The direct propagation approach is shown in Algorithm~\ref{alg:directprop}. This approach performs one forward and one backward propagation for each acquired feature, so the computational complexity is $O(m)$.

\begin{table}
\begin{tabular}{c|c}
\begin{minipage}[t]{0.48\textwidth}
\begin{algorithm}[H]
   \caption{Direct propagation algorithm}
   \label{alg:directprop}
\begin{algorithmic}
   \STATE {\bfseries Input:} Feature vector $\textbf{x}_{partial}$, model $M$, cost vector $\textbf{c}$
   \STATE $ t \gets 0, \mathbf{k} \gets \mathbf{0} $
   \STATE $ \textbf{x}_{\mathit{filled}}^t \gets \textbf{k} \odot \textbf{x}_{\mathit{partial}} + (\textbf{1}-\textbf{k}) \odot \mathbb{E}[\textbf{x}] $
   \REPEAT
    \STATE $ \hat{\textbf{y}} \gets M.\mathit{predict}(\textbf{x}^t_{\mathit{filled}}) $
    \STATE $ \textbf{r} \gets \mathit{abs}(\mathit{get\_relevances}(\textbf{x}_{\mathit{filled}}^t, \hat{\textbf{y}})) $
    \STATE $ \hat{\textbf{r}} \gets (\textbf{r} \odot (\mathbf{1} - \textbf{k})) \oslash \textbf{c} $
  
    \STATE $ \mathit{next\_feature} \gets \mathit{argmax}_i (\hat{\textbf{r}}) $
    \STATE $\textbf{x}_{\mathit{filled}}^{t+1} \gets \mathit{acquire} (\textbf{x}^t_{\mathit{filled}}, \mathit{next\_feature}) $
    \STATE $ k_{\mathit{next\_feature}} \gets 1 $
    \STATE $ t \gets t+1 $
   \UNTIL{Stopping condition is reached}
   \STATE {\bfseries return} $M.predict(\textbf{x}^t_{filled})$
\end{algorithmic}
\end{algorithm}
\end{minipage}
&
\begin{minipage}[t]{0.48\textwidth}
\begin{algorithm}[H]
   \caption{Multiple propagations algorithm}
   \label{alg:multiprop}
    \begin{algorithmic}
        \STATE {\bfseries Input:} Feature vector $\textbf{x}_{\mathit{partial}}$, model $M$, cost vector $\textbf{c}$
        \STATE $ t \gets 0, \mathbf{k} \gets \mathbf{0} $
        \STATE $ \textbf{x}_{\mathit{filled}} \gets \textbf{k} \odot \textbf{x}_{\mathit{partial}} + (\textbf{1}-\textbf{k}) \odot \mathbb{E}[\textbf{x}] $
        \REPEAT
        \STATE $ \hat{y} \gets M.\mathit{predict}(\textbf{x}^t_{\mathit{filled}}) $
        \STATE $ \mathit{best\_relevance} \gets -\infty $
        \STATE $ \mathit{best\_feature} \gets 0 $
        \FOR{$i \leftarrow 0$ \ldots $\mathit{n\_classes}$}
            \STATE $\textbf{relevance\_out} \gets \mathbf{0} $
            \STATE $\mathit{relevance\_out}_i \gets 1 $
            \STATE $ \textbf{r} \gets \mathit{get\_relevances}(\textbf{x}_{\mathit{filled}}^t, \textbf{relevance\_out}) $
            \STATE $ \hat{\textbf{r}} \gets (\mathit{abs}(\textbf{r}) \odot (\mathbf{1} - \textbf{k})) \oslash \textbf{c} $
            \IF{ $max(\hat{\textbf{r}}) > \mathit{best\_relevance}$ }
                \STATE $ \mathit{best\_feature} \gets \mathit{argmax}_i    (\hat{\textbf{r}}) $
                \STATE $ \mathit{best\_relevance} \gets max(\hat{\textbf{r}}) $
            \ENDIF
            
        \ENDFOR
        \STATE $ \textbf{x}_{\mathit{filled}}^{t+1} \gets \mathit{acquire} (\textbf{x}^t_{\mathit{filled}}, \mathit{best\_feature}) $
        \STATE $ \textbf{k}_{\mathit{best\_feature}} \gets 1 $
        \STATE $ t \gets t+1 $
        \UNTIL{Stopping condition is reached}
       \STATE {\bfseries return} $M.predict(\textbf{x}^t_{filled})$
    \end{algorithmic}
\end{algorithm}
\end{minipage}
\end{tabular}
\end{table}

\subsection{Multiple Propagations}
Our second algorithm modifies the direct propagation algorithm to take into account relevant features for each output node $\hat{y}_i$. The intuition behind this approach is that some output nodes might receive a low predicted value due to some unknown input feature, which can cause that input feature to have a low relevance in the direct propagation algorithm. By propagating relevance through each output node separately, we can avoid such a situation at the cost of increased computational complexity.

The problem setting is the same as in our first algorithm. However, when the relevance was propagated backward in the previous method, this time we set $\hat{y}_i = 1$ for one $i$, while $\hat{y}_j = 0, \forall j \neq i$. This is then propagated backward using the same rules as previously. This process is repeated for all values of $i$, so in total the backward propagation is performed $n$ times, once for each output class. We then choose the globally maximal adjusted relevance $\hat{r}_i$, and acquire the value of feature $i$.

The multiple propagations method is shown in Algorithm~\ref{alg:multiprop}. This method is computationally more demanding than the first method, as each feature acquisition requires one forward pass and $n$ backward passes, thereby making the computational complexity $O(nm)$. However, the algorithm can be modified to perform the backward passes in parallel, as they do not depend on each other. For clarity, only the serial method is shown here.

%


\section{Results}

We evaluate our approach on three cost-sensitive datasets: diabetes prediction, heart disease prediction, and Yahoo! Learning to Rank Competition dataset (LTRC). Summarization of these datasets can be found in Table~\ref{table:datasets}.

\begin{table*}[ht]
\centering
\caption{Datasets and their corresponding neural networks}
\begin{tabular}{ p{4.6em} p{4.6em} p{4.6em} p{4.6em} p{4.6em} p{4.6em} } 
\hline
\textbf{Dataset} & \textbf{Examples} & \textbf{Features} & \textbf{Classes} & \textbf{Hidden layers} & \textbf{Layer size} \\
\hline
LTRC & 30000 & 500 & 5 & 2 & 32 \\ 
Diabetes & 25474 & 581 & 3 & 2 & 128 \\ 
Heart & 49509 & 245 & 2 & 1 & 32 \\ 
\hline
\end{tabular}
\label{table:datasets}
\end{table*}

\subsection{Diabetes Prediction}

As medical diagnosis is an area where active feature acquisition can be highly beneficial, the first two experiments use medical datasets. We have derived two datasets from The National Health and Nutrition Examination Survey (NHANES), which is a long-term program that has collected health and nutrition data from a nationally representative group of 5000 people between years 1999--2016 \cite{cdc_nhanes}. The full dataset contains questionnaire answers as well as results from physical and laboratory examinations. As this dataset contains data on a wide range of medical problems, we will first focus on predicting whether an individual has diabetes, prediabetes, or no diabetes.

To predict the level of diabetes, we look at the blood glucose levels and define an individual as healthy if their fasting plasma glucose level is below 100~mg/dL, prediabetic if the level is between 100 and 125~mg/dL, and diabetic if the level is over 125~mg/dL. These ranges have been defined by \citet{cdc_2017}. We filter out features that are directly related to our target variable, i.e. any variable measuring blood glucose, as they would make the prediction trivial. We also leave out variables that have missing values for over 25\% of the subjects. We define feature costs based on a rough estimate on how much money and effort is needed to acquire that feature. The feature costs are listed in Table~\ref{table:nhanes_cost}. The dataset is provided in the supplemental materials with the code to reproduce our results. The data was split into a training set (70\%) and a test set (30\%) randomly. Both sets were balanced by oversampling to have an equal number of examples from each class. All features were normalized to range $[0,1]$.

\begin{table*}[ht]
\centering
\caption{Feature costs for NHANES dataset}
\label{table:nhanes_cost}
\begin{tabular}{ p{10em} p{2em} } 
\hline
\textbf{Feature type} & \textbf{Cost} \\
\hline
Demographics & 1 \\ 
Questionnaire answer & 5 \\ 
Physical examination & 10 \\ 
Laboratory test & 100 \\ 
\hline
\end{tabular}
\end{table*}

\begin{figure}[htb]
\centering
\begin{subfigure}[t]{0.49\textwidth}
    \centering
    \includegraphics[width=\textwidth]{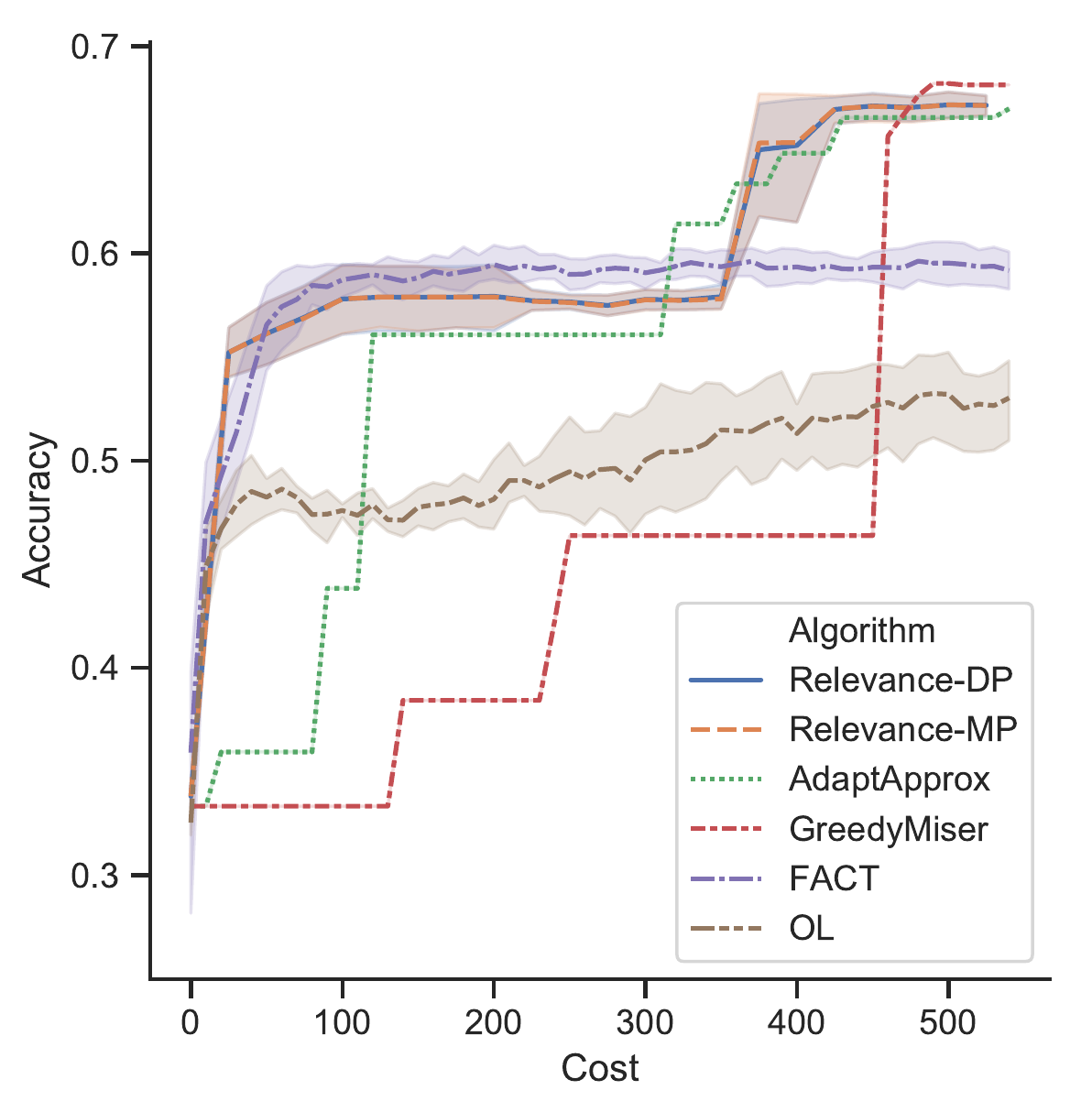}
    \caption{Diabetes}
    \label{fig:nhanes}
\end{subfigure}
\hfill
\begin{subfigure}[t]{0.49\textwidth}
    \centering
    \includegraphics[width=\textwidth]{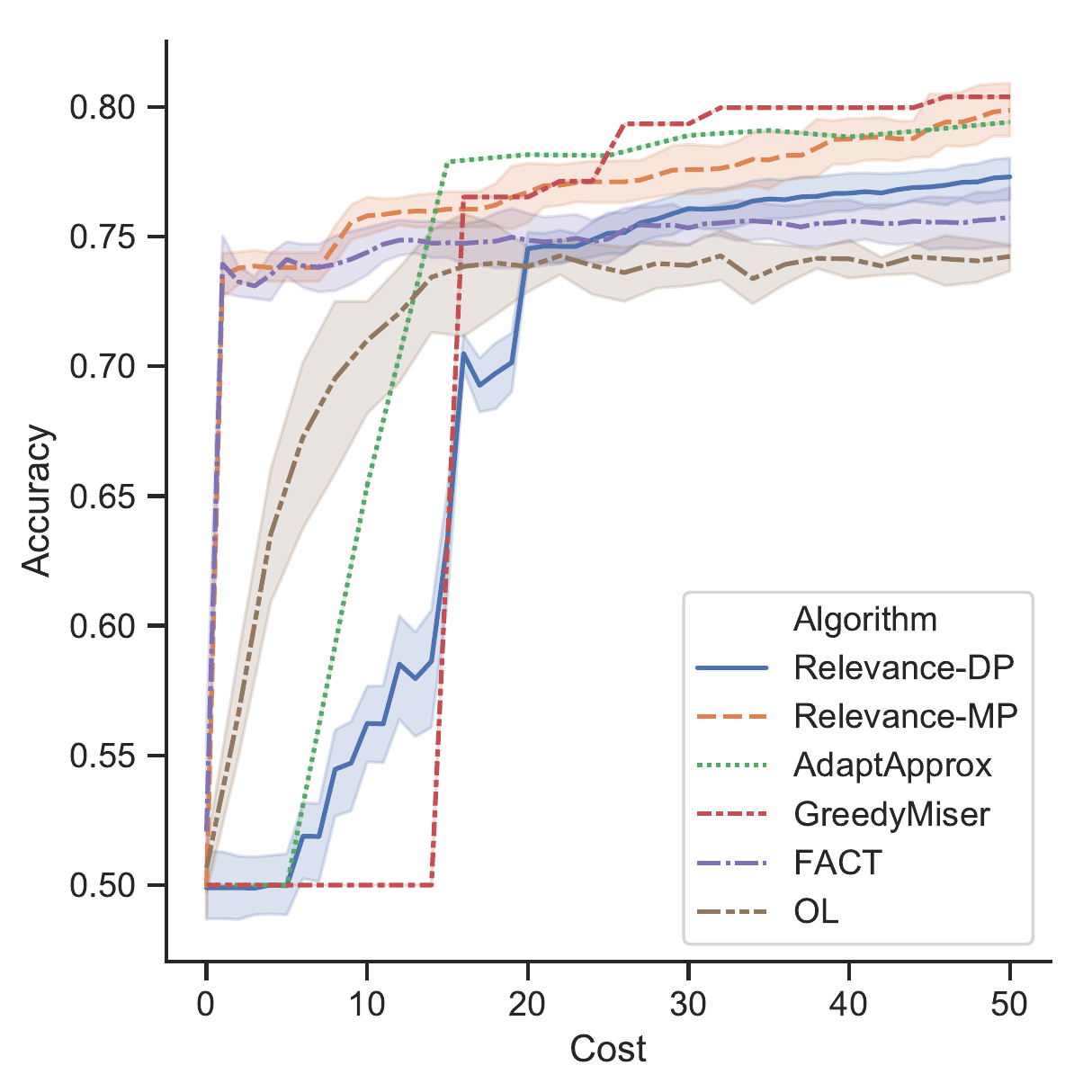}
    \caption{Heart disease}
    \label{fig:heart}
\end{subfigure}
\hfill
\caption{Prediction accuracy on (a) diabetes and (b) heart disease dataset.}
\end{figure}

We compare our approach to four other algorithms: The Greedy Miser \cite{Xu2012TheBudgets}, AdaptApprox \cite{nan2017adaptive}, FACT \cite{kachuee2018dynamic} and Opportunistic Learning (OL) \cite{kachuee2019opportunistic}. We selected these algorithms based on their good performance and the availability of reference implementations. The Greedy Miser is an algorithm for learning cost-sensitive classification and regression trees. AdaptApprox learns a gating function to decide whether to use a cheap or an expensive classifier. FACT uses neural network's sensitivity to determine which feature is most likely to change the prediction. Opportunistic Learning uses reinforcement learning to learn what features to ask. The results of the first experiment can be found in Figure~\ref{fig:nhanes}. 

In this experiment, both of our proposed algorithms (Relevance-DP, Relevance-MP) provide nearly identical results. Initially, FACT provides similar results to our algorithms, but converges to a lower accuracy. Opportunistic Learning (OL) improves its accuracy very slowly after the initial features. AdaptApprox reaches a good accuracy but slightly slower than our algorithms. The Greedy Miser has to acquire a large number of features before finally achieving similar accuracy to the other algorithms. In this experiment, our algorithms provide a fast convergence and a high final accuracy, thereby combining the best aspects of the other algorithms.

\subsection{Heart Disease Prediction}

Our next dataset is also derived from the NHANES dataset \cite{cdc_nhanes}. This time the goal is to predict whether an individual has a heart disease, such as congestive heart failure, or has had a heart attack. We use the same costs and data preprocessing steps as in the previous dataset, shown in Table~\ref{table:nhanes_cost}. The heart disease dataset can be found in the supplemental materials with the code that can be used to reproduce our results.

The results are shown in Figure~\ref{fig:heart}. This time there is a significant difference between our proposed algorithms. Using multiple propagations (Relevance-MP) provides faster convergence than using direct propagation (Relevance-DP). Again, FACT provides similar results to our Relevance-MP algorithm initially but converges to a lower accuracy. Opportunistic Learning (OL) starts slightly slower and converges to a similar accuracy with FACT. AdaptApprox starts with more expensive features, therefore staying at a low accuracy for longer. The Greedy Miser suffers from the same problem, but also reaches a good accuracy later. This experiment shows the benefit of using multiple propagations instead of the more simple algorithm. 

\subsection{Learning to Rank Competition}

The goal of Yahoo! Learning to Rank Competition (LTRC) dataset is to predict how relevant a specific document is given a query \cite{Chapelle2011YahooOverview}. The relevance is defined by an expert using a five-step scale. The feature vectors consist of features describing the query and the document. These features include e.g. how many times a document has been clicked on the result list, how recent the page is, how well the document's text matches the query, and so on. Each feature has an associated feature extraction cost between 1--200, which has been defined by Yahoo!. The split into training and evaluation sets has been defined in the original dataset and was used as-is for our experiments.

\citet{Chapelle2011YahooOverview} suggests using Normalized Discounted Cumulative Gain (NDCG) to measure the quality of the predictions. This metric was introduced by \citet{Jarvelin2002CumulatedTechniques}, and it compares the relevance of predicted result order to the optimal order. It has also been used by previous feature acquisition papers \cite{Xu2012TheBudgets,xu2013cost}, so we will use it to measure the performance of our algorithms as well.

As the ranking problem is different from the traditional classification problem, we compare our results to the algorithms that have been designed for this problem and have demonstrated the best performance: Cost-Sensitive Tree of Classifiers (CSTC) \cite{xu2013cost}, Cronus \cite{Chen2012ClassifierCost} and Early Exit \cite{cambazoglu2010early}. CSTC builds a tree of classifiers, where the chosen path determines which features will be used for prediction. Cronus builds a cascade of classifiers so that the easy inputs will be handled by the earlier classifiers in the cascade with a low cost, and the more complicated inputs will go through more classifiers leading to a higher cost. Early Exit scores documents gradually, dropping out the ones with too low scores before fully evaluating them. 

\begin{figure}[htb]
\centering
\begin{subfigure}[t]{0.49\textwidth}
    \centering
    \includegraphics[width=\textwidth]{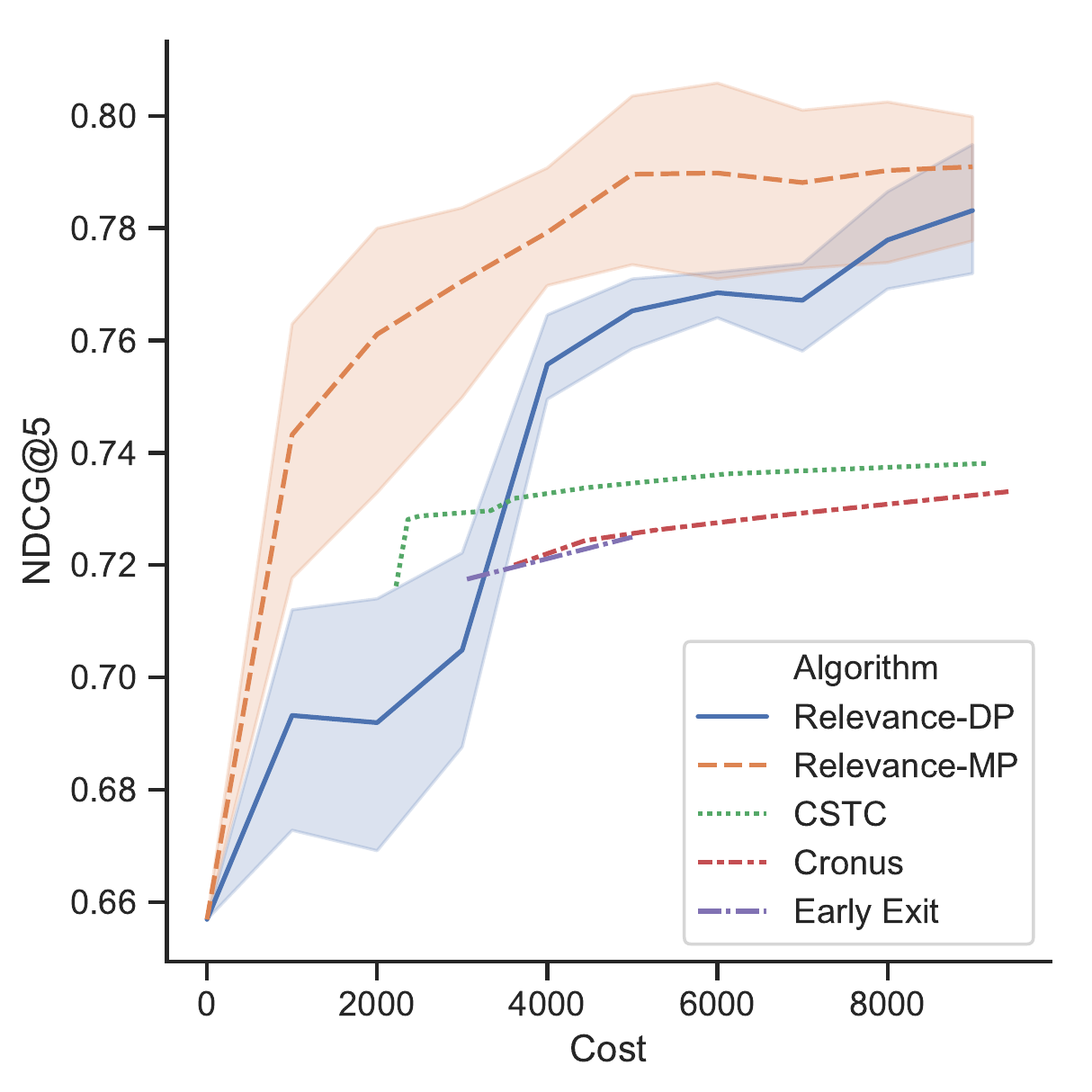}
\end{subfigure}
\caption{NDCG@5 score on Yahoo Learning to Rank Competition dataset.}
\label{fig:ltrc}
\end{figure}

The results are shown in Figure~\ref{fig:ltrc}. As can be seen, feature acquisition with relevance propagation reaches a higher NDCG@5 score than the other approaches. In addition, using multiple propagations (Relevance-MP) converges to a high score significantly faster than using the direct propagation algorithm (Relevance-DP). 

\section{Conclusion}

In this paper, we have shown a novel approach for the cost-sensitive active feature acquisition problem. Our approach uses missing feature relevance as the core idea for choosing which features to acquire. We have demonstrated two different propagation algorithms using this approach. First of them provided good results with one forward and one backward propagation per acquired feature, while the second algorithm improved the results further at the expense of a slightly increased computational cost. We evaluated the proposed algorithms on three realistic datasets: Yahoo! Learning to Rank Competition dataset and two health datasets derived from National Health and Nutrition Examination Survey (NHANES) dataset. Our results show that our first algorithm (direct propagation) performs well in most cases, while our second algorithm (multiple propagations) is more robust and out-performs the existing algorithms.

\bibliographystyle{plainnat}
\bibliography{biblio2,bibliography}

\end{document}